\documentclass[10pt,twocolumn]{article}

\usepackage[utf8]{inputenc}

\usepackage[T1]{fontenc}

\usepackage{amsmath,amssymb,amsthm}

\usepackage{graphicx}

\usepackage{multirow}
\usepackage{tabularx}
\usepackage{array}

\usepackage{booktabs}

\usepackage{algorithm}

\usepackage{algorithmic}

\usepackage{xcolor}

\usepackage[colorlinks=true,linkcolor=blue,citecolor=blue,urlcolor=blue]{hyperref}

\usepackage[numbers]{natbib}
\title{Culture-Aware Humorous Captioning: Multimodal Humor Generation across Cultural Contexts}

\author{Run Xu$^{1}$, Lu Li$^{2}$, Rongzhao Zhang$^{3,\dagger}$, Jie Xu$^{3,\dagger}$\\
$^{1}$Nanyang Technological University, \texttt{run001@e.ntu.edu.sg}\\
$^{2}$Tongji University, \texttt{author2@yyy.edu}\\
$^{3}$Shanghai Artificial Intelligence Laboratory, \texttt{zhangrongzhao@pjlab.org.cn, xujie@pjlab.org.cn}\\
}

\date{}

\begin{document}
\maketitle

\begin{figure*}[t]
  \centering
  \includegraphics[width=\textwidth]{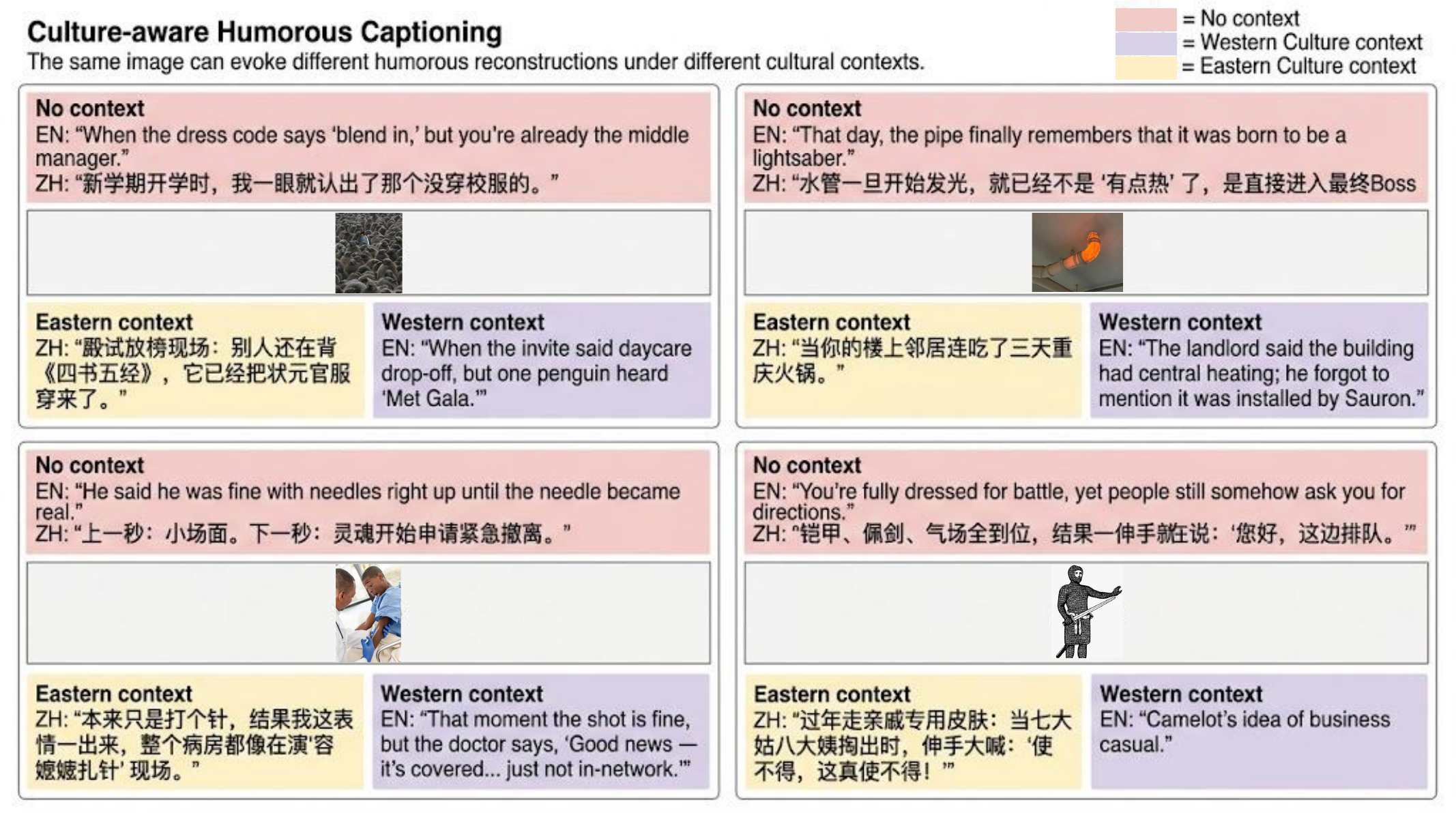}
  \caption{Qualitative examples of culture-aware humorous captioning under different culture contexts. The same image can trigger distinct humorous captions with different framing and cultural references when conditioned on no context, Western context, or Eastern context.}
  \label{fig:teaser}
\end{figure*}

\begin{abstract}
Recent multimodal large language models have shown promising ability in generating humorous captions for images, yet they still lack stable control over explicit cultural context, making it difficult to jointly maintain image relevance, contextual appropriateness, and humor quality under a specified cultural background. To address this limitation, we introduce a new multimodal generation task, culture-aware humorous captioning, which requires a model to generate a humorous caption conditioned on both an input image and a target cultural context. Captions generated under different cultural contexts are not expected to share the same surface form, but should remain grounded in similar visual situations or humorous rationales.

To support this task, we establish a six-dimensional evaluation framework covering image relevance, contextual fit, semantic richness, reasonableness, humor, and creativity. We further propose a staged alignment framework that first initializes the model with high-resource supervision under the Western cultural context, then performs multi-dimensional preference alignment via judge-based GRPO with a Degradation-aware Prototype Repulsion Constraint to mitigate reward hacking in open-ended generation, and finally adapts the model to the Eastern cultural context with a small amount of supervision.

Experimental results show that our method achieves stronger overall performance under the proposed evaluation framework, with particularly large gains in contextual fit and a better balance between image relevance and humor under cultural constraints.
\end{abstract}

\section{Introduction}
Multimodal large language models (MLLMs) have recently shown strong progress in visual understanding and open-ended text generation, but they still remain unreliable on generation tasks that require fine-grained visual grounding, explicit controllability, and complex pragmatic objectives \cite{liu2024improved,yuan2024osprey, peng2025patch,lu2025benchmarking,lee2024fleur,cheng2025caparena,lian2025describe,chen2025compcap,zhang2025sc,tu2025ode,lee2025diffusion}.This limitation becomes more evident in humorous caption generation, where the model must go beyond literal description and coordinate visual cues, implicit meaning, and expressive novelty \cite{let,nandy2024yesbut,saakyan2025understanding,ryan2025humor,humordb,hwang2025bottlehumor}. At the same time, recent studies on LLMs and MLLMs have increasingly shown that culture is not a superficial label, but a factor that systematically shapes commonsense activation, expression preferences, and task completion strategies \cite{shen2024understanding,adilazuarda2024towards,investigating,culturellm,li2024culturepark,bhatt2024extrinsic,liu2025culturally,maji2025drishtikon,kabir2025break,wu2025incorporating,wu2025socialcc,guo2025care}. These observations motivate a new question: whether a model can generate humorous captions that remain visually grounded while adapting to an explicitly specified cultural context.

However, existing work has not systematically modeled the intersection of humor and culture in multimodal generation. Prior multimodal humor studies mainly focus on understanding or generating punchlines, sarcasm, and metaphor \cite{let,nandy2024yesbut,saakyan2025understanding,ryan2025humor,humordb,hwang2025bottlehumor}, while cultural studies primarily examine cultural commonsense, value alignment, and open-ended cultural evaluation in LLMs and MLLMs \cite{shen2024understanding,adilazuarda2024towards,investigating,culturellm,li2024culturepark,bhatt2024extrinsic,liu2025culturally,maji2025drishtikon,kabir2025break,wu2025incorporating,wu2025socialcc,guo2025care}. As a result, there is still no clear task formulation, evaluation protocol, or alignment method for generating humorous captions under explicit cultural context. The key difficulty is that the same image may evoke different shared experiences, default assumptions, and associative pathways across cultures, leading to different yet individually reasonable humorous expressions. Therefore, cultural context should be treated not as a surface style tag, but as a condition that directly affects how humor is constructed from the image \cite{let,nandy2024yesbut,saakyan2025understanding,ryan2025humor,humordb,hwang2025bottlehumor,shen2024understanding,adilazuarda2024towards,culturellm,li2024culturepark,bhatt2024extrinsic,liu2025culturally,maji2025drishtikon,kabir2025break,wu2025incorporating,wu2025socialcc,guo2025care}.

To address this gap, we propose culture-aware humorous captioning, a task in which the model generates a humorous caption conditioned on both an input image and a target cultural context. Compared with general humorous captioning, this task requires the model to satisfy three objectives simultaneously: preserving stable grounding in the image, making the target cultural context genuinely participate in humor construction, and still producing a caption with a perceptible punchline and expressive novelty. This also makes evaluation more difficult, since a single metric is insufficient once the task becomes open-ended, culturally conditioned, and humor-sensitive \cite{peng2025patch,lu2025benchmarking,lee2024fleur,cheng2025caparena,lian2025describe,chen2025compcap,zhang2025sc,tu2025ode,lee2025diffusion,lee2025diffusion}. We therefore formulate the task under coarse-grained Eastern and Western cultural context settings and build a six-dimensional evaluation framework covering Image Relevance, Contextual Fit, Semantic Richness, Reasonableness, Humor, and Creativity.

Based on this formulation, we propose a staged alignment framework for culture-aware humorous captioning. The model is first initialized with high-resource supervision under the Western cultural context, then optimized through multi-dimensional preference alignment via judge-based GRPO, and finally adapted to the Eastern cultural context with a small amount of supervision. To reduce reward hacking in open-ended preference optimization, we further introduce a Degradation-aware Prototype Repulsion Constraint. Experimental results show that this framework more effectively converts explicit cultural conditions into actual generation capability, yielding especially clear gains in Contextual Fit while better balancing image relevance and humor under cultural constraints.

\section{Related Works}
\subsection{Humorous Image Captioning}
Humorous image captioning aims to generate captions that are not only grounded in the visual content of an image but also convey a recognizable humorous effect. Humorous image captioning has progressed from early supervised joke generation to large-scale, reasoning-driven, and preference-aware multi-modal systems. Recent methods can be roughly grouped into two lines: benchmark-building efforts that scale humorous captioning into a systematic research problem \cite{Oxford,humorinai,memecap,humordb,punmemecn}, and  reasoning- or preference-based models that improve specificity, creativity, or alignment with human judgments \cite{bridging,let,xmecap,humorchain,wings}. Despite significant advancements, most prior work optimizes funniness, content-specificity, or pairwise preference under implicitly shared social assumptions, rather than requiring stable responses aligned with explicit specified cultural context. In response, our work formulates and explores the potential of culture-aware humorous captioning, where humor must remain image-grounded while adapting its rationale and expression to the target culture.

\subsection{Controllable Image Captioning and Culture-Aware Generation}
Recent controllable captioning studies can be summarized into content or structure control \cite{learning,controlcap,mcoca}, context-aware control \cite{controllable,repic}, and style-aware control \cite{visual, captionsmiths, anycap}. In parallel, recent work on cultural alignment in LLMs suggests that explicit cultural conditioning can substantially reshape model behavior and often yield responses that are better aligned with culture-specific values, norms, and situational expectations \cite{investigating,culturellm,cultural}. However, these two lines of work remain weakly connected: controllable captioning usually emphasizes regions, attributes, or surface style, whereas culture-aware generation is largely text-only and rarely grounded in images or humor. Our task therefore goes beyond style transfer by treating cultural context as a deeper interpretive condition that changes what knowledge is activated, which incongruity is foregrounded, and how humor is verbalized.

\subsection{Evaluation and Alignment of MLLMs}
Current evaluation of MLLMs mainly follows two paradigms: benchmark-based testing for standardized measurement of multimodal perception and reasoning, and judge-based assessment for open-ended generation. On the benchmark side, recent datasets provide broad and systematic evaluation of multimodal capabilities \cite{mmbench,seed,mmmu}. For open-ended outputs, evaluation has increasingly adopted LLM-as-a-Judge and MLLM-as-a-Judge frameworks as scalable evaluators\cite{judgelm,llava,generation,mllm,judging,judge,crowd,evaluating}. In parallel, alignment of MLLMs is typically achieved through supervised fine-tuning, RLHF-style optimization, or preference-based post-training \cite{mammoth,rlhf,multi,mm,bi2025llava,commit,repic,task,re}. For creative captioning in particular, large-scale human preference data has proven highly effective for both evaluation and alignment \cite{humorinai}. However, these general paradigms remain insufficient or partly inapplicable for culture-aware humorous captioning, where a valid output should jointly satisfy image grounding, contextual fit, semantic richness, coherence, humor, and creativity. This mismatch motivates our task-specific six-dimensional evaluation framework and staged alignment strategy tailored to open-ended, culture-conditioned humorous generation.

\section{Task Definition}
Humorous caption generation is not determined solely by the explicit visual content of an image, but is also substantially influenced by external context. For the same visual situation, different cultural backgrounds may activate different shared experiences, default assumptions, and associative pathways, thereby changing how the image is interpreted and how humor is constructed. Therefore, generating humorous captions under a cultural constraint is not a matter of adding superficial stylistic variation to a general humorous caption. Instead, it requires the model to produce different yet reasonable humorous reconstructions of the same visual situation, conditioned on the target cultural context while remaining grounded in the image.

Based on this observation, we define the task of culture-aware humorous captioning. Given an input image $x$ and a text condition $c$ that explicitly specifies the target cultural context, the model is required to generate a short, natural, and humorous caption $y$. The generation process can be formulated as:
\begin{equation}
\label{eq:conditional-generation}
y \sim p(y \mid x, c)
\end{equation}
where $c$ indicates the target cultural background, for example, generating a humorous caption under the Eastern cultural context or the Western cultural context.

In this work, cultural context is instantiated using two coarse-grained conditions: the Eastern cultural context and the Western cultural context. This distinction is introduced for task modeling purposes, with the goal of capturing broad differences in associative patterns and expressive tendencies across cultural backgrounds, rather than providing a fine-grained or exhaustive definition of culture.

Unlike conventional image captioning, the goal of our task is not to provide an objective transcription of image content. Instead, the model must organize a humorous caption for the same visual situation by combining visual understanding with the shared experiences, expressive conventions, and potential associative pathways associated with the target cultural context. Accordingly, captions generated for the same image under different cultural contexts are not expected to share the same surface wording or even exactly the same humorous trigger. More importantly, however, they should all remain centered on the same input image and maintain clear semantic connections to the entities, actions, scenes, or relations depicted in it. In other words, differences across cultural versions should primarily lie in the interpretive pathway and humor construction, rather than in arbitrary divergence from the image content.

From the perspective of task objectives, an ideal output should satisfy at least three requirements. First, the caption should maintain stable correspondence with the visible content of the image, rather than introducing core information without visual support. Second, it should reflect more natural associative patterns and expressive logic under the target cultural context, rather than creating superficial variation by mechanically inserting cultural symbols. Third, the caption itself should exhibit a perceptible humorous effect while remaining concise. Based on this task definition, we next establish a six-dimensional evaluation framework to characterize generation quality in a more systematic manner.

\section{Evaluation Framework}
Generating humorous captions under cultural constraints simultaneously involves visual understanding, contextual alignment, linguistic organization, and humorous expression, making it difficult to characterize output quality with a single metric. A caption may be highly relevant to the image yet fail to reflect the target cultural context, or it may align with the cultural context while lacking a clear humorous effect. To enable a more stable and fine-grained analysis of model performance on this task, we establish a six-dimensional evaluation framework covering Image Relevance, Contextual Fit, Semantic Richness, Reasonableness, Humor, and Creativity.

Overall, these six dimensions can be grouped into three categories. The first category concerns visual-context alignment, including Image Relevance and Contextual Fit. The second concerns semantic and expressive quality, including Semantic Richness and Reasonableness. The third concerns humor performance, including Humor and Creativity. The first two categories are used to evaluate whether the output is genuinely grounded in both the image and the target cultural context, while the last category assesses whether the caption further develops effective and non-trivial humorous expression on that basis.

Specifically, Image Relevance (IR) measures whether the caption is anchored in the visible content of the image. Contextual Fit (CF) measures whether the caption naturally reflects the associative patterns and expressive logic of the target cultural context. Semantic Richness (SR) measures whether the text provides additional semantic layers or interpretive space beyond surface-level description. Reasonableness (Ra) measures whether the expression is natural, coherent, and broadly consistent with commonsense. Humor (Hu) measures whether the output contains a perceptible humorous effect. Creativity (Cr) measures whether the idea or expression shows novelty and avoids overly templated responses.

To improve evaluation consistency and interpretability, each dimension is scored on a 10-point scale, together with a four-level rubric constraining different score ranges. Specifically, scores of 0–2 indicate extremely poor performance on the target dimension, typically involving severe distortion, obvious conflict, or near-complete failure. Scores of 3–5 indicate relatively weak performance: the dimension is only partially satisfied and still exhibits clear deficiencies. Scores of 6–7 indicate generally good performance: the major requirements are met, but there remains room for improvement in completeness, detail, or stability. Scores of 8–10 indicate strong performance with high completion quality and only minor or negligible defects. For this task, scores in the 8–10 range usually correspond to relatively strong and comparatively rare high-quality outputs. We report both the score of each individual dimension and their overall average to present model performance from multiple perspectives (See Appendix B for the full evaluation protocol).

\section{Data Construction}
To support culture-aware humorous captioning, we construct both training/development data and a cross-cultural benchmark under a unified task definition: generating a short humorous caption conditioned on an input image and a target cultural context. Starting from the raw source pool, we retain 5,000 task-suitable images and divide them into three disjoint subsets: 3,500 for training, 500 for development, and 1,000 for benchmark evaluation. Table \ref{tab:data_stats} summarizes the resulting data organization.

\begin{table}[t]

\centering

\caption{Dataset organization. The three splits are image-disjoint. ``Asym.'' denotes asymmetric cultural organization for training, and ``Sym.'' denotes symmetric dual-context references for evaluation.}
\label{tab:data_stats}

\small

\setlength{\tabcolsep}{4pt}

\begin{tabularx}{\columnwidth}{l c c c >{\raggedright\arraybackslash}X}

\toprule

Split & Images & Role & Culture & Output \\

\midrule

Train & 3500 & Training   & Asym. & Captions + refusal \\

Dev   & 500  & Validation & Asym. & Captions \\

Benchmark & 1000 & Evaluation & Sym. & Western/Eastern refs. \\

\bottomrule

\end{tabularx}

\end{table}

The raw images are collected from CLoT. We filter them according to visual richness and reinterpretability, favoring samples that contain object relations, action interactions, scene cues, or situational contrast that can support culture-conditioned humorous reconstruction. We remove text-dominant images and multi-panel composites to avoid heavy OCR dependence and cross-panel narrative complexity. Explicit cultural symbols are treated as a preference rather than a strict requirement. The final filtered pool contains 5,000 images.

All high-quality captioned data in this work are constructed under a unified pipeline consisting of candidate generation, automatic screening, and manual spot-checking (see the appendix for details), with additional review of ambiguous cases when necessary. This pipeline is applied consistently across both training supervision and benchmark reference construction. Different subsets mainly differ in their functional role and in the model used for candidate generation, rather than in the quality-control procedure itself.

For the training data, we adopt an asymmetric cross-cultural design. Large-scale captions under the Western cultural context are used to learn shared capabilities, including visual grounding, conditional response, and humor organization, while captions under the Eastern cultural context are reserved as a smaller high-quality resource for later adaptation. We additionally include a small number of refusal samples to model context-mismatch cases, where the image does not provide sufficient support for the target cultural context. For standard caption samples, candidate captions are generated for each image-context pair and one caption is retained after screening; for refusal samples, a unified short template is used as the target output. Quality control focuses on image relevance, contextual fit, linguistic naturalness, and humor effectiveness.

The benchmark is constructed on the 1,000-image test split, which is fully disjoint from the training and development sets. For each benchmark image, we build reference captions under both the Western cultural context and the Eastern cultural context using the same high-quality construction pipeline. These references are not intended to exhaust all valid humorous expressions; instead, they provide stable and high-quality anchors for evaluating Image Relevance, Contextual Fit, Humor, and Creativity under controlled cross-cultural comparison.

\begin{figure*}[!t]
  \centering
  \includegraphics[width=\textwidth]{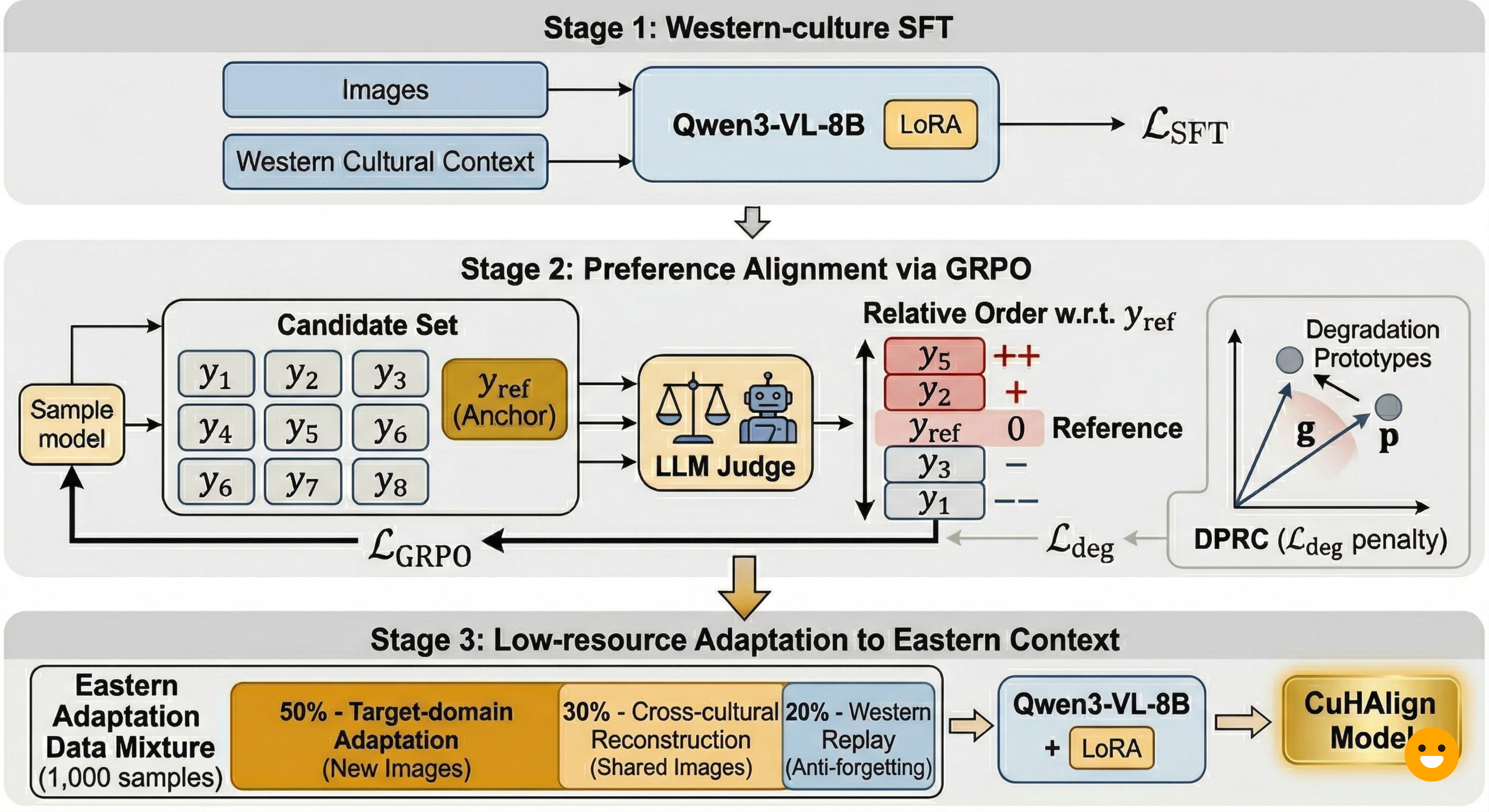}
  \caption{Overall three-stage framework of CuHAlign. We first perform SFT on a Western-culture dataset to initialize the task, training the model to generate humorous captions from the input image and target cultural context (Stage 1). We then conduct preference alignment with a judge-based GRPO trainer, where multiple rollouts are ranked against a reference under a cultural-humor rubric and jointly optimized with a degradation-aware prototype repulsion constraint (Stage 2). Finally, the aligned model is adapted to the Eastern cultural context through few-shot SFT on the Eastern dataset, while a small-size replay set of Western dataset is jointly used to mitigate forgetting (Stage 3).}
  \label{fig:framework}
\end{figure*}

\section{Method}
\subsection{Method Overview}

For the task of culture-aware humorous captioning, we propose CuHAlign, a staged alignment framework, as in Figure \ref{fig:framework}. Given an input image $x$ and a target cultural context condition $c$, the model generates a caption $y \sim p(y \mid x, c)$. Since this task requires the model to jointly satisfy Image Relevance, Contextual Fit, and Humor, a single supervision objective is insufficient to fully optimize generation quality. We therefore adopt a three-stage training pipeline. First, the model is initialized with high-resource supervision under the Western cultural context to establish task response patterns and output format constraints. Second, multi-dimensional preference alignment is performed via judge-based GRPO, together with a Degradation-aware Prototype Repulsion Constraint to mitigate reward hacking in open-ended generation. Third, the model is adapted with a small amount of supervision under the Eastern cultural context to support low-resource cultural transfer. This design is consistent with the data construction strategy in Section 5: large-scale data under the Western cultural context is used to learn shared capabilities, while smaller-scale data under the Eastern cultural context is reserved for later adaptation.

\subsection{Backbone Model and Task Initialization}
We adopt Qwen3-VL-8B \cite{bai2025qwen3} as the backbone model and use LoRA for parameter-efficient fine-tuning. Preliminary analysis suggests that, although the backbone already possesses strong visual understanding ability, it remains insufficient in responding to explicit cultural conditions, producing compressed humorous captions, and controlling output format. To bring the model distribution into the target task space, in the first stage we conduct supervised fine-tuning on 3,000 image-text samples under the Western cultural context from the training pool described in Section 5. Given a training sample $(x_i, c_i, y_i)$, the optimization objective is
\begin{equation}
 \mathcal{L}_{\mathrm{SFT}} = - \sum_i \log p_\theta(y_i \mid x_i, c_i).
\end{equation}
This stage mainly establishes instruction following for the task, caption-format constraints, and the basic ability to generate humorous captions under a cultural context. However, supervised fine-tuning alone cannot directly optimize the joint balance among Image Relevance, Contextual Fit, and Humor, which motivates the subsequent preference alignment stage.

\subsection{Multi-dimensional Preference Alignment via an LLM Judge}
Since our task is an open-ended creative generation problem without an explicitly verifiable reward, we use an LLM-as-a-Judge to provide preference signals. For each training sample, the current policy model samples $K=8$ rollouts, which, together with the reference caption, form a candidate set. Given the input image, the cultural context condition, the candidate captions, and the evaluation rubric, the judge outputs a joint ranking over the candidate set. Compared with directly predicting absolute scores, within-group ranking can partially alleviate the instability of the judge’s scoring scale.

We treat the reference caption as an anchor and construct rewards only for the sampled rollouts. Let $\mathrm{rank}_j$ denote the rank of the j-th rollout in the joint ranking, and let $\mathrm{rank}_{\mathrm{ref}}$ denote the rank of the reference caption. The raw reward of rollout $j$ is defined as
\begin{equation}
r_j = f(\mathrm{rank}_j, \mathrm{rank}_{\mathrm{ref}}),
\end{equation}
where $f(\cdot)$ maps the relative rank difference to a bounded scalar: rollouts ranked above the reference receive positive rewards, while those ranked below it receive negative rewards, with the magnitude increasing as the rank gap grows. We then perform a relative-advantage transformation on $\{r_j\}_{j=1}^{K}$ within each sampled group, and use the resulting advantages for GRPO updates.

After obtaining within-group relative rewards, we optimize the policy using GRPO. Unlike supervised learning, which performs token-level likelihood fitting to a single reference answer, GRPO still operates on the token probability distribution of the generated sequence, but weights it according to sequence-level relative advantage. As a result, the optimization target shifts from “reproducing the reference caption” to “increasing the probability of outputs that better satisfy multi-dimensional preferences.”

To alleviate reward hacking in judge-based preference optimization, we further introduce a Degradation-aware Prototype Repulsion Constraint. For training samples with degradation annotations (see the appendix for details), each annotation contains one or more degradation directions together with their corresponding evidence texts. Before training, we aggregate evidence texts by degradation direction and map them into a shared representation space using a unified text encoder. The mean representation of evidence texts from the same direction is taken as the prototype vector $\mathbf{p}$ for that degradation direction. For a generated output, we use the same encoder to obtain its sequence-level representation $\mathbf{g}$. For generated samples with degradation annotations, we compute the cosine similarity between 
$\mathbf{g}$ and the corresponding degradation prototype. When the similarity exceeds a threshold m, a repulsion penalty is imposed:
\begin{equation}
    \mathcal{L}_{\mathrm{deg}} = \max\bigl(0, \cos(\mathbf{g}, \mathbf{p}) - m\bigr).
\end{equation}
If a sample is associated with multiple degradation directions, the losses for different directions are combined by weighted summation to obtain $\mathcal{L}_{\mathrm{deg}}$. The final optimization objective is
\begin{equation}
    \mathcal{L} = \mathcal{L}_{\mathrm{GRPO}} + \lambda \mathcal{L}_{\mathrm{deg}}.
\end{equation}
Here, $\mathcal{L}_{\mathrm{deg}}$ is activated only for samples with degradation annotations, serving as an additional structured negative constraint around known degradation neighborhoods. It complements the judge-ranking signal and reduces the risk that the policy exploits local rubric preferences while drifting away from the true task objective.

\subsection{Low-resource Adaptation to the Eastern Cultural Context}
After the first two stages, the model has acquired relatively strong shared capabilities under the high-resource distribution of the Western cultural context, including visual understanding, conditional response, caption organization, and multi-dimensional preference balancing. However, because these capabilities are mainly learned from high-resource Western-distribution data, the model’s default associative pathways and expressive tendencies may still remain biased toward that context. To address this issue, in the final stage we introduce a lightweight adaptation process using data under the Eastern cultural context to calibrate the model toward the low-resource target setting.

This stage uses 1,000 image-text pairs in total. Among them, $50\%$ are new images paired with captions under the Eastern cultural context, which are used for target-domain adaptation; $30\%$ are Eastern-context captions for images already appearing in the training pool, which are used to learn cross-cultural humorous reconstruction for the same visual situation; and the remaining $20\%$ are supervised samples under the Western cultural context, which are replayed to mitigate catastrophic forgetting. The training procedure is the same as in the first stage and still uses supervised fine-tuning. However, the goal here is no longer to establish basic task ability, but to perform targeted adaptation of the shared capabilities to the low-resource Eastern cultural context.

\section{Experiments}
\subsection{Experimental Setup}
We construct a cross-cultural benchmark containing 1,000 images in total . Each test sample consists of an input image and an explicitly specified cultural context condition, and the model is required to generate a short humorous caption. To ensure fair comparison, all models are evaluated on the same test split, and results are reported separately under the Western cultural context and the Eastern cultural context.

To reduce additional bias caused by cross-lingual mismatch, the Eastern cultural context setting uses Chinese prompts for inference, Chinese prompts for evaluation, and Chinese generated outputs, while the Western cultural context  setting uses English prompts for inference, English prompts for evaluation, and English generated outputs. Accordingly, our analysis mainly focuses on relative model performance within the same cultural setting, as well as the difference between the same model with and without explicit cultural conditioning.

We adopt the six-dimensional evaluation framework introduced in Section 4, including Image Relevance (IR), Contextual Fit (CF), Semantic Richness (SR), Reasonableness (Ra), Humor (Hu), and Creativity (Cr). Each dimension is scored on a 10-point scale, and the overall score is computed as the unweighted average of the six dimensions. Evaluation follows a human-machine hybrid protocol: among the 1,000 test samples, a fixed $20\%$ subset is scored by human annotators, while the remaining $80\%$ is scored automatically by an LLM-as-a-Judge under a unified rubric. For each metric, the final result is obtained by directly averaging the human-scored portion and the judge-scored portion over the full test set, thereby combining manual reliability with scalable automatic evaluation. The compared baselines include general-purpose multimodal large language models, the Qwen3-VL-8B backbone, and the prior relevant method CLoT\cite{let}. In all subsequent experiments, we report both the six individual dimension scores and the overall average, with particular attention to CF, since it most directly reflects the model’s ability to respond to explicit cultural context conditions.Additional qualitative examples and failure cases are provided in Appendix.

\subsection{Main Results}

\begin{table*}[t]
\centering
\caption{Results under Western and Eastern culture contexts.}
\label{tab:main_results}
\footnotesize
\setlength{\tabcolsep}{3pt}
\resizebox{\textwidth}{!}{
\begin{tabular}{lcccccccccccccc}
\toprule
& \multicolumn{7}{c}{Western Culture Context} & \multicolumn{7}{c}{Eastern Culture Context} \\
\cmidrule(lr){2-8} \cmidrule(lr){9-15}
Model & IR & CF & SR & Ra & Hu & Cr & Avg. & IR & CF & SR & Ra & Hu & Cr & Avg. \\
\midrule
GPT-4o \cite{hurst2024gpt} & 7.27 & 6.11 & 5.91 & 8.06 & 6.76 & 6.47 & 6.76 & 7.51 & 5.72 & 5.96 & 8.04 & 6.74 & 6.50 & 6.75 \\
gemini-3-flash-preview \cite{google2026gemini3flashpreview} & 7.32 & 5.71 & 5.86 & 8.14 & 6.92 & 6.25 & 6.70 & 7.23 & 5.23 & 5.89 & 8.01 & 6.83 & 6.17 & 6.56 \\
Claude Sonnet 4.5\_pred \cite{anthropic2025claudesonnet45systemcard} & 7.11 & 5.66 & 5.89 & 8.05 & 6.82 & 6.21 & 6.62 & 7.21 & 5.11 & 5.72 & 7.97 & 6.78 & 6.16 & 6.49 \\
\midrule
InternVL3-8B-Instruct \cite{zhu2025internvl3} & 7.08 & 4.90 & 4.95 & 7.79 & 6.00 & 5.61 & 6.06 & 6.83 & 4.79 & 5.24 & 7.78 & 6.10 & 5.78 & 6.09 \\
LLaVA-OneVision-7B \cite{li2024llava} & 6.02 & 4.67 & 4.44 & 6.97 & 4.55 & 4.64 & 5.21 & 6.03 & 3.15 & 3.40 & 6.60 & 3.69 & 3.74 & 4.44 \\
MiniCPM-V 2.6 \cite{yao2024minicpm} & 6.82 & 5.17 & 5.21 & 7.75 & 5.86 & 5.82 & 6.10 & 6.52 & 3.80 & 4.37 & 7.15 & 4.87 & 4.61 & 5.22 \\
MiniGPT-4 \cite{zhu2023minigpt} & 6.54 & 3.94 & 2.95 & 6.98 & 3.97 & 3.36 & 4.62 & 4.88 & 2.41 & 2.52 & 5.73 & 1.90 & 2.32 & 3.29 \\
GLM-4V-9B \cite{glm2024chatglm} & 4.79 & 3.70 & 3.60 & 5.67 & 2.25 & 3.01 & 3.84 & 5.42 & 3.01 & 3.37 & 6.62 & 2.72 & 3.14 & 4.05 \\
Qwen2.5-VL-7B-Instruct \cite{Bai2025Qwen25VLTR} & 6.60 & 4.94 & 4.93 & 7.61 & 5.92 & 5.55 & 5.92 & 6.88 & 4.19 & 4.87 & 7.74 & 5.39 & 5.31 & 5.73 \\
Qwen3-VL-8B-Instruct \cite{bai2025qwen3} & 7.04 & 5.49 & 5.61 & 7.84 & 6.41 & 6.20 & 6.43 & 7.17 & 5.22 & 5.73 & 7.87 & 6.63 & 6.28 & 6.48 \\
cogvlm2-llama3-chat-19B \cite{hong2024cogvlm2} & 6.92 & 4.93 & 5.03 & 7.76 & 5.80 & 5.57 & 6.00 & 7.20 & 3.93 & 4.52 & 7.73 & 5.02 & 4.84 & 5.54 \\
\midrule
QwenVL+CLoT \cite{let} & 7.11 & 4.32 & 4.38 & 7.80 & 5.67 & 5.02 & 5.72 & 7.26 & 3.75 & 3.56 & 7.76 & 4.80 & 4.10 & 5.20 \\
CuHAlign & 6.88 & 6.63 & 5.60 & 7.84 & 6.67 & 6.16 & 6.63 & 6.89 & 6.30 & 5.91 & 7.88 & 6.68 & 6.30 & 6.66 \\
\bottomrule
\end{tabular}
}
\end{table*}

Tables \ref{tab:main_results} report the main results under the Western cultural context and the Eastern cultural context, respectively. Overall, existing multimodal large language models already exhibit a certain capability for humorous image captioning. However, under explicit cultural constraints, performance differences across models are much larger on Contextual Fit (CF) than on several dimensions more closely related to general caption quality. This indicates that generating an amusing caption and generating a humorous caption that is appropriate for a target cultural context are not the same capability; the latter places substantially higher demands on conditional response and contextual modeling.

In contrast, CuHAlign achieves the best CF under both cultural settings while maintaining strong overall quality. Under the Western cultural context, CuHAlign obtains an overall average score of 6.63 and a CF score of 6.63. Under the Eastern cultural context, it achieves an overall average of 6.66 and a CF score of 6.30. Compared with the Qwen3-VL-8B backbone, CuHAlign improves CF from 5.49 to 6.63 under the Western setting and from 5.22 to 6.30 under the Eastern setting, showing that the proposed method more effectively translates explicit cultural conditions into actual generation capability. At the same time, CuHAlign remains highly competitive in overall average score under both settings, indicating that these gains are not obtained by sacrificing general generation quality.

More importantly, the advantage of CuHAlign is concentrated on the task-specific dimension of Contextual Fit, rather than being merely a uniform improvement in general caption quality. This result is consistent with both the task formulation and the method design: for culture-aware humorous captioning, the key is not simply to generate more generally amusing text, but to make the target cultural context genuinely participate in humor construction while preserving relevance to the image. The main results therefore support the necessity of studying culture-aware humorous captioning as an independent research problem and demonstrate that the proposed staged alignment framework can serve this goal effectively and stably.

\subsection{Effect of Explicit Cultural Context}

\begin{table*}[t]
\centering
\caption{Effect of explicit cultural context conditioning.}
\label{tab:context_effect}
\begin{tabular}{ccccccc}
\toprule
\multirow{2}{*}{Model} 
& \multicolumn{2}{c}{No Context} 
& \multicolumn{2}{c}{Western Context} 
& \multicolumn{2}{c}{Eastern Context} \\
\cmidrule(lr){2-3} \cmidrule(lr){4-5} \cmidrule(lr){6-7}
& CF & Overall & CF & Overall & CF & Overall \\
\midrule
InternVL3-8B-Instruct \cite{zhu2025internvl3} & 3.55 & 5.84 & 4.90 & 6.06 & 4.79 & 6.09 \\
MiniCPM-V 2.6 \cite{yao2024minicpm}         & 3.34 & 5.43 & 5.17 & 6.10 & 3.80 & 5.22 \\
Qwen3-VL-8B-Instruct \cite{bai2025qwen3}  & 3.97 & 6.24 & 5.49 & 6.43 & 5.22 & 6.48 \\
CuHAlign              & 5.21 & 6.37 & 6.63 & 6.63 & 6.30 & 6.66 \\
\bottomrule
\end{tabular}
\end{table*}

\begin{table*}[t]
\centering
\caption{Ablation study of the staged alignment framework under Western and Eastern cultural settings. The best results are highlighted in bold.}
\label{tab:ablation_main}
\footnotesize
\setlength{\tabcolsep}{3.5pt}
\resizebox{\textwidth}{!}{
\begin{tabular}{lccccccccccccc}
\toprule
\multirow{2}{*}{Model Variant} & \multirow{2}{*}{SFT} & \multirow{2}{*}{GRPO} & \multirow{2}{*}{Deg.} & \multirow{2}{*}{E.\ Adapt.} 
& \multicolumn{4}{c}{Western Context} & \multicolumn{4}{c}{Eastern Context} \\
\cmidrule(lr){6-9} \cmidrule(lr){10-13}
& & & & & IR & CF & Hu & Overall & IR & CF & Hu & Overall \\
\midrule
Base                  &   &   &   &   & \textbf{7.04} & 5.49 & 6.41 & 6.43 & 7.17 & 5.22 & 6.63 & 6.48 \\
+ SFT                 & \checkmark &   &   &   & 6.42 & 6.13 & 6.25 & 6.38 & 6.80 & 5.25 & 5.98 & 6.01 \\
+ GRPO                & \checkmark & \checkmark &   &   & 6.86 & 6.59 & 6.47 & 6.60 & 6.84 & 5.40 & 5.97 & 6.07 \\
+ Deg.                & \checkmark & \checkmark & \checkmark &   & 6.88 & 6.62 & 6.65 & 6.61 & 6.87 & 5.43 & 5.97 & 6.09 \\
CuHAlign (full model) & \checkmark & \checkmark & \checkmark & \checkmark & 6.88 & \textbf{6.63} & \textbf{6.67} & \textbf{6.63} & \textbf{6.89} & \textbf{6.30} & \textbf{6.68} & \textbf{6.66} \\
\bottomrule
\end{tabular}
}
\end{table*}

To verify the necessity of explicit cultural conditioning in this task, we compare model performance under three settings: the Western cultural context, the Eastern cultural context, and no cultural context. All results are obtained on the same 1,000 test samples and are evaluated under the unified six-dimensional framework introduced in Section 4.

Overall, as in Table \ref{tab:context_effect}, the introduction of explicit cultural context leads to the most significant gains on Contextual Fit (CF) across models. For example, for Qwen3-VL-8B, the CF score increases from 3.97 in the no-context setting to 5.49 and 5.22 under the Western and Eastern settings, respectively. Its overall average score also rises from 6.24 to 6.43 and 6.48. By comparison, CuHAlign makes more effective use of explicit cultural conditions: its CF score increases from 5.21 in the no-context setting to 6.63 and 6.30 under the Western and Eastern settings, while its overall average score improves from 6.37 to 6.63 and 6.66.

These results show that explicit cultural context is not an optional auxiliary prompt, but a factor that substantially changes how the model interprets the image and organizes humorous expression. Meanwhile, the larger CF gains achieved by CuHAlign under culturally conditioned settings indicate that the proposed method can not only generate humorous captions in a general sense, but can also more reliably convert the target cultural context into actual generation behavior. This finding is consistent with the main results in Section 7.2 and the ablation analysis in Section 7.4, and further supports the necessity of distinguishing culture-aware humorous captioning from general humorous captioning.

\subsection{Ablation Study of the Staged Alignment Framework}

To examine the contribution of each component in the staged alignment framework proposed in Section 6, we conduct ablation studies under both the Western cultural context and the Eastern cultural context. The compared variants are, in order: the backbone model Qwen3-VL-8B; the model after supervised fine-tuning under the Western cultural context; the model further enhanced with GRPO; and the model further augmented with the Degradation-aware Prototype Repulsion Constraint. On top of these variants, we then add the final low-resource adaptation stage under the Eastern cultural context, together with a small amount of replayed Western supervision to mitigate catastrophic forgetting, yielding the final CuHAlign full model. Results are shown in Table \ref{tab:ablation_main}. 

Under the Western cultural context, the ablation results show a clear progressive trend. Compared with the backbone, Western-context supervised fine-tuning improves CF from 5.49 to 6.13, indicating that the first stage effectively establishes task response behavior and output format. After adding GRPO, the overall average score further increases to 6.60, and CF rises to 6.59, showing that multi-dimensional preference alignment is the key step for improving overall quality. Adding the Degradation-aware Prototype Repulsion Constraint further improves the overall average to 6.61, suggesting that this constraint provides additional benefit in balancing open-ended generation quality. Finally, the full model achieves the best results in this group, with an overall average of 6.63 and a CF score of 6.63.

Under the Eastern cultural context, the role of the final adaptation stage is even more pronounced. After Western-context supervised fine-tuning, GRPO, and the degradation-aware constraint, CF improves progressively from 5.25 to 5.40 and 5.43. On this basis, after further adaptation to the Eastern cultural context, the full model reaches an overall average of 6.66 and a CF score of 6.30, while Humor (Hu) and Creativity (Cr) also increase to 6.68 and 6.30, respectively. These results indicate that the first two stages mainly learn transferable task structure and alignment capability, whereas the final lightweight cultural adaptation stage effectively calibrates these shared capabilities toward the low-resource Eastern setting.

Overall, the ablation results support the functional division of CuHAlign. The first-stage supervised fine-tuning under the Western cultural context mainly establishes task format and basic conditional response. The second-stage GRPO is the most important step for improving generation quality. The Degradation-aware Prototype Repulsion Constraint further improves the overall balance in open-ended generation. The final adaptation stage under the Eastern cultural context is crucial for improving performance in the low-resource cultural setting. Notably, the full model achieves the highest CF under both cultural settings, further confirming that the staged framework can stably transform explicit cultural conditions into actual generation capability.

\subsection{Judge Reliability and Agreement with Human Evaluation}

To assess the reliability of automatic evaluation, we further construct a dedicated judge validation set. Specifically, we sample 400 image instances from the test set and build pairwise evaluation samples of the form image + high-quality caption + low-quality caption (see the appendix for details). The high-quality captions are taken from the human reference captions in the dataset, while the low-quality captions are mainly sampled from relatively poor outputs of baseline models on the corresponding images, with an effort to cover different types of quality defects.

\begin{figure}[t]
  \centering
  \includegraphics[width=\linewidth]{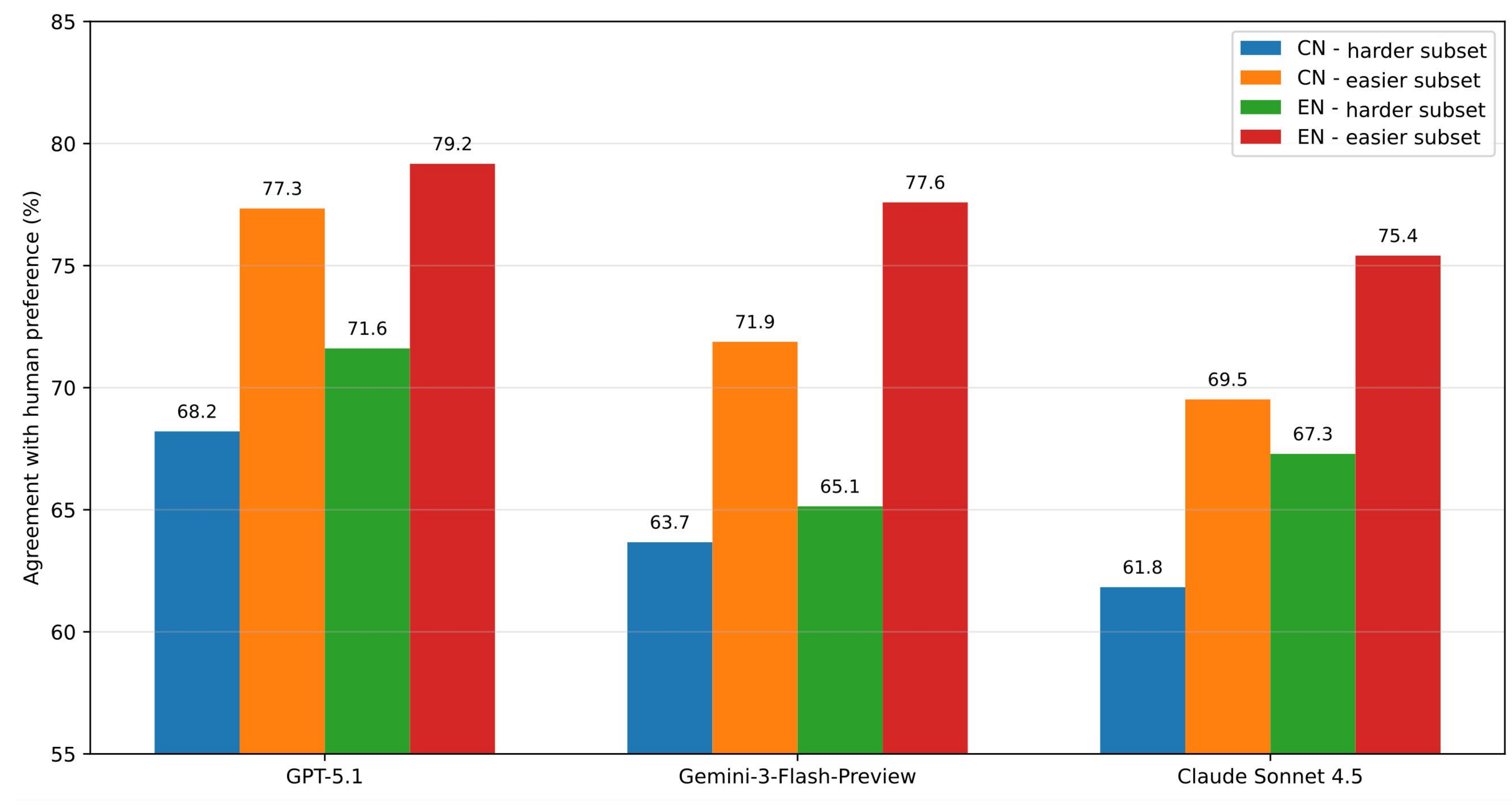}
  \caption{Comparison of three LLM judges on the judge validation set for assessing the reliability of automatic pairwise evaluation across languages and difficulty levels.}
  \label{fig:judgeresults}

\end{figure}

To distinguish judge performance under different levels of difficulty, we further divide the validation set into two subsets. The first 200 cases contain caption pairs with relatively large quality gaps, while the remaining 200 contain pairs with smaller quality differences and more ambiguous decision boundaries. We then ask GPT-5.1, Gemini-3-Flash-Preview\cite{google2026gemini3flashpreview}, and Claude Sonnet 4.5\cite{anthropic2025claudesonnet45systemcard} to perform pairwise judgments on these samples, and compute their agreement rates with the manually predefined better-worse labels. The results are reported in Figure \ref{fig:judgeresults}.

Overall, all three judges achieve higher agreement on samples with clear quality differences, while accuracy drops on more ambiguous pairs. This indicates that automatic evaluation is reliable for distinguishing outputs with obvious quality gaps, but remains challenging for fine-grained discrimination. Our validation focuses on pairwise comparison rather than absolute score calibration, since the latter is inherently difficult for culture-aware humorous captioning due to its open-ended and multi-dimensional nature. In practice, the judge in our framework mainly provides relative quality signals, so strong better-worse discrimination already offers meaningful evidence of judge adequacy for both evaluation and preference alignment. Among the three judges, GPT-5.1 achieves the highest agreement in all four settings and shows the most stable overall performance, so we adopt a hybrid protocol with $20\%$ human scoring and $80\%$judge scoring to balance scalability and reliability.

\section{Conclusion}
We introduce culture-aware humorous captioning and propose CuHAlign to improve multimodal humor generation under explicit cultural contexts. Results show clear gains, especially in contextual fit, while the current Chinese–Eastern and English–Western setup may not fully disentangle language effects from cultural effects.


\bibliography{references}

\end{document}